\documentclass[letterpaper, 10 pt, conference]{ieeeconf}  

\IEEEoverridecommandlockouts                              
\usepackage[l2tabu,orthodox]{nag}

\usepackage[
    backend=bibtex8,
    style=ieee,
    sorting=none,
    natbib=true,
    doi=false,
    isbn=false,
    url=false,
    eprint=false,
    maxcitenames=1,
    mincitenames=1
]{biblatex}

\DefineBibliographyStrings{english}{
    andothers = {\textit{et\addabbrvspace al\adddot}},
}

\usepackage[pdftex,colorlinks]{hyperref}

\usepackage[printonlyused]{acronym}

\usepackage{siunitx}
\usepackage[all]{nowidow}

\usepackage[dvipsnames]{xcolor}

\usepackage{lipsum}

\usepackage{xspace} 
\newcommand{\ie}{\textit{i.e.},\;\xspace{}}
\newcommand{\eg}{\textit{e.g.},\;\xspace{}}

\usepackage[pdftex]{graphicx}

\usepackage{epstopdf}

\usepackage{import}

\graphicspath{{./latexGoodPractices/}}

\usepackage{booktabs}

\usepackage{tabularx}
\usepackage{multirow, multicol}

\usepackage{amssymb,amsfonts,amsmath,amscd}

\usepackage{bm}

\newcommand{\bbm}{\begin{bmatrix}}
\newcommand{\ebm}{\end{bmatrix}}

\usepackage{physics}  
\usepackage{amsmath} 
\usepackage[normalem]{ulem}
\usepackage{booktabs}    
\usepackage{tabularx}    
\usepackage{pifont}      
\usepackage{amssymb}     
\usepackage[dvipsnames]{xcolor}
\usepackage[capitalize, nameinlink]{cleveref} 

\newcommand{\SE}{\mathrm{{SE}}}

\title{\LARGE \bfseries 
Comparing Trajectories from Positions Alone: \\ Curvature-Based Time Alignment and Drift Error Metric
} 

\author{Effie Daum $^{1}$, Daniele De Martini $^{2}$, Claire Dune $^{1}$, François Pomerleau $^{1}$
\thanks{$^{1}$Northern Robotics Laboratory, Université Laval, Québec City, Québec, Canada
		{\texttt{\small effie.daum@norlab.ulaval.ca}} and \texttt{\small {francois.pomerleau@ift.ulaval.ca}}}%
\thanks{$^{2}$Mobile Robotics Group, Oxford Robotics Institute, University of Oxford, Oxford, United Kingdom}
\thanks{*This work was supported by the Canada Research Chair on Field Robotics (CRC-2023-00396) and the NSERC (ALLRP 585289-23) project CRYOTIC.} 
}

\usepackage[switch]{lineno}
\usepackage[font=small]{caption}
\usepackage{subcaption}

\acrodef{ATE}{Absolute Trajectory Error}
\acrodef{APTE}{Absolute Palindrome Trajectory Error}
\acrodef{DoF}{Degrees-of-Freedom}
\acrodef{DOP}{Dilution Of Precision}
\acrodef{DLIO}{Direct Lidar Inertial Odometry}
\acrodef{D-GNSS}{Differential-GNSS}
\acrodef{DE}{Drift Error}
\acrodef{EKF}{Extended Kalman Filter}
\acrodef{FOG}{Fiber-Optic Gyroscope}
\acrodef{ICR}{Instant Center of Rotation}
\acrodef{IMU}{Inertial Measurement Unit}
\acrodef{ICP}{Iterative Closest Point}
\acrodef{INS}{Inertial Navigation System}
\acrodef{GNSS}{Global Navigation Satellite System}
\acrodef{GTF}{Ground Truth Free}
\acrodef{HDOP}{Horizontal Dilution Of Precision}
\acrodef{LOS}{Line of Sight}
\acrodef{mAA}{mean Average Accuracy}
\acrodef{NLOS}{Non-Line of Sight}
\acrodef{NTP}{Network Time Protocol}
\acrodef{PAS}{Pose Alignment Score}
\acrodef{PLT}{Positioning Layout System}
\acrodef{PPK}{Post-Processed Kinematic}
\acrodef{PPP}{Precise Point Positioning}
\acrodef{PTP}{Precision Time Protocol}
\acrodef{RAS}{Rotation Alignment Score}
\acrodef{RMSE}{Root Mean Square Error}
\acrodef{RPE}{Relative Pose Error}
\acrodef{RTK}{Real-Time Kinematic}
\acrodef{RTS}{Robotic Total Station}
\acrodef{SfM}{Structure-from-Motion}
\acrodef{SLAM}{Simultaneous Localization And Mapping}
\acrodef{TAS}{Translation Alignment Score}
\acrodef{TLS}{Terrestrial Laser Scanner}
\acrodef{UGV}{Uncrewed Ground Vehicule}
\acrodef{WILN}{Weather Invariant Lidar Navigation}
\acrodef{VO}{Visual Odometry}
\acrodef{LO}{Lidar Odometry}
 \acrodef{IQR}{Interquartile Range}

\begin{document}
\maketitle
\thispagestyle{empty}
\pagestyle{empty}

\begin{abstract}
In field robotics, acquiring independent large-scale reference trajectories more accurate than the evaluated estimates remains an open challenge.
The domain is widely reliant on \ac{ATE} and \ac{RPE}, computed with automated tools, that rest on assumptions and evaluation parameters rarely made explicit.
When unreported, the errors can be misleading and hinder fair comparisons. 
This paper introduces a trajectory-evaluation protocol for standardized and reliable accuracy assessment in state estimation, localization, and \ac{SLAM}. 
The approach combines a novel temporal alignment method based on curvature signals with an error metric normalized by travelled distance.
We explicitly account for temporal synchronization, sampling alignment, and extrinsic calibration, quantifying their influence through a sensitivity analysis. 
The proposed protocol contributes to more rigorous, reproducible, and standardized trajectory evaluation.
\end{abstract}


\acresetall 

\section{Introduction}
\label{sec:introduction}
Claims of progress in robotics must be supported by rigorous evaluation and benchmarking, whether in perception, manipulation, locomotion, or navigation~\citep{bonsignorio2015toward}. 
Ideally, such benchmarking should rely on publicly available datasets that provide all necessary input data, including timestamped sensor measurements and intrinsic and extrinsic calibration parameters. 
The data streams are assumed to be accurately synchronized, and their respective reference frames to be consistently defined. 
Central to any benchmark is a trustworthy reference (\ie often referred to as ground truth): an independent set of measurements whose accuracy exceeds that of the method under evaluation by a sufficient margin~\citep{ceriani2009rawseeds}.
In practice, however, achieving such conditions is particularly challenging in field environments: reference measurements are affected by noise, sensor clocks may drift, and calibration parameters inevitably carry some uncertainty. 
In particular, spatial and temporal misalignments between estimated and reference trajectories can substantially affect reported error values~\citep{frey2025boxi}, yet their influence is frequently overlooked in the literature~\citep{zhang2019rethinking}. 
Rather than requiring flawless data acquisition, explicitly accounting for these imperfections within the evaluation protocol would provide access to a much broader range of existing datasets suitable for rigorous evaluation.

\begin{figure}[t]
	\centering
	\includegraphics[width=\columnwidth]{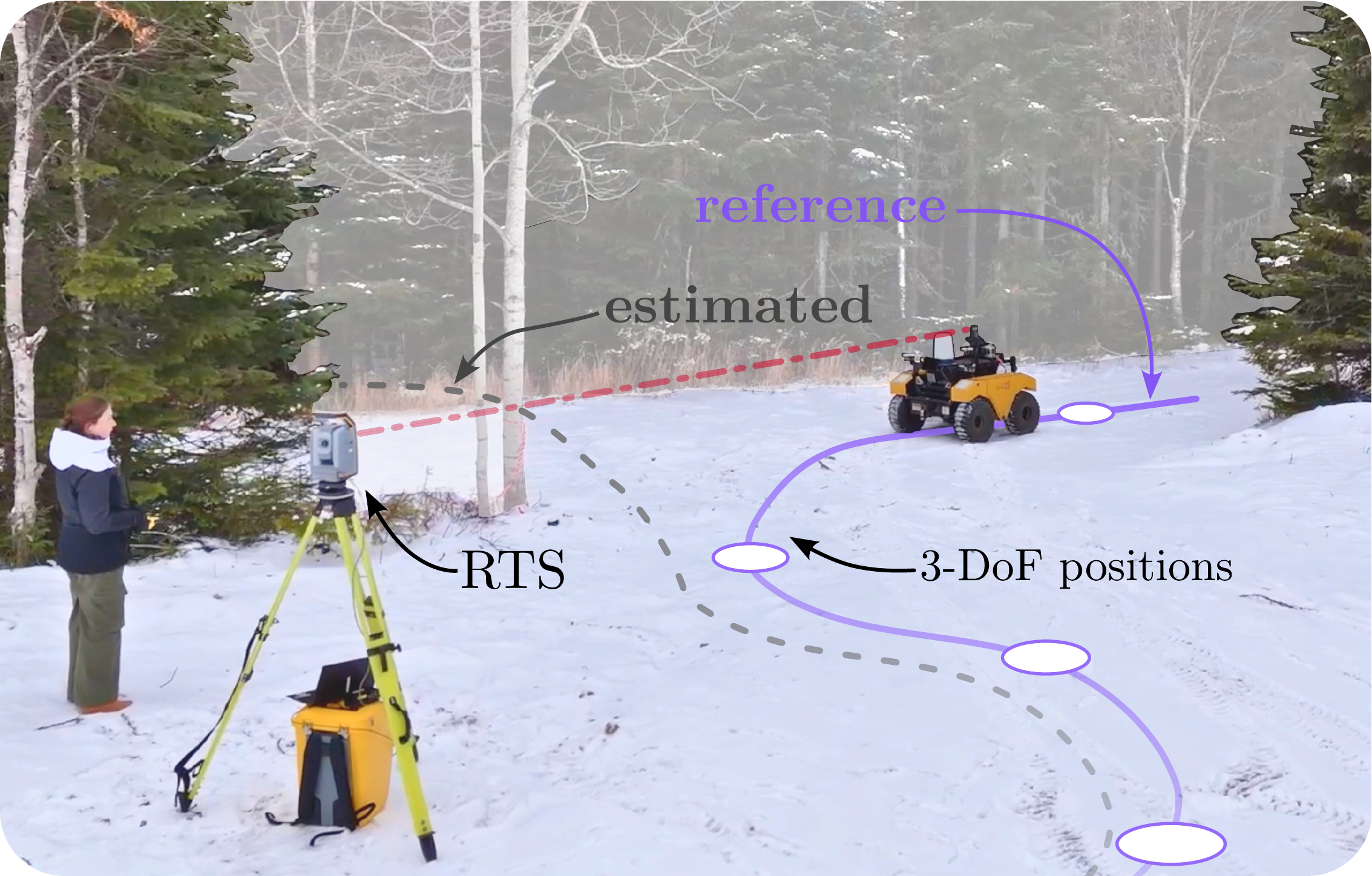}
	\caption{
    Limitations of real-world experiments in capturing reference trajectories in a forest.
    The reference trajectory is shown in purple, with white ellipses indicating positions collected with a \acs{RTS} or \acs{GNSS}, \ie without orientation.
    The estimated trajectory is represented in dashed gray, gathered with a state estimation algorithm by the robotic platform.
    Our proposed evaluation protocol allows us to evaluate the estimated trajectory in these conditions, where traditional metrics such as \ac{RPE} are undefined.
    We propose to temporally align trajectories and evaluate errors using our \ac{DE} metric. 
    }
    \label{fig:intro}
    \vspace{-0.21in}
\end{figure}

Currently, two metrics are widely used for trajectory benchmarking: \ac{ATE} and \acf{RPE}~\citep{sturm2012benchmark}. 
However, their reported values depend on evaluation parameters that are rarely documented consistently, making comparisons across studies potentially unreliable~\citep{frey2026grandtour}. 
\ac{ATE} depends on the global alignment between estimated and reference trajectories, which may absorb part of the estimation error.
\ac{RPE} requires full 6-\ac{DoF} reference poses~\citep{burri2016euroc}, whereas large-scale outdoor reference systems such as \ac{GNSS} or \ac{RTS} typically provide only 3D positions, as illustrated in~\cref{fig:intro}. 
Reconstructing the missing orientations using additional sensors, such as an \ac{IMU}~\citep{mittelstedt2025factor}, compromises ground-truth independence when the same sensors contribute to the evaluated estimator~\citep{ceriani2009rawseeds}.
Instead of proposing a workaround for this limitation, we offer a profound analysis of metrics integrated in a global evaluation protocol.

Therefore, in this work, we evaluate trajectories from 3D positions alone, without known frame alignment, using curvature signature. 
Curvature can be extracted from positions, making it well-suited and invariant to unknown trajectory orientations.
Our contributions are:
\begin{enumerate}
\renewcommand{\labelenumi}{\emph{(\roman{enumi})}}
\item A temporal alignment method matching trajectory curvature signatures.
\item A formal introduction to \acf{DE}, a trajectory error metric expressed with respect to the distance travelled.
\item An evaluation protocol combining temporal alignment, spatial registration, and \ac{DE}, validated on the GrandTour~\citep{frey2026grandtour} and the FoMo~\citep{boxan2026fomo} datasets.
\end{enumerate}

\section{Related Work}
\label{sec:related_work}
\subsection{Establishing a common time base}
\label{sec:RW-C1}
Evaluating a trajectory against a reference assumes that both are expressed in a common time base, which must be established either at acquisition time or at evaluation time.
In the trajectory evaluation protocols, the clocks are assumed to be already synchronized, so that establishing the temporal correspondence reduces to an association step~\citep{zhang2019rethinking}.
Accordingly, evaluation toolkits match the two trajectories by nearest timestamps within a duration threshold~\citep{zhang2018tutorial}, a sufficient method for experimental setups with high recording frequency and where synchronization is assumed to be handled at acquisition time.
For recordings where these assumptions are broken, \citet{kelly2014general} propose a time calibration algorithm cast as a registration problem with Time-Delay~\ac{ICP}.
Formulated on curves in orientation space, the method starts from a coarse initial alignment and alternates two steps until convergence: each sample is matched to its closest counterpart on the other curve, and the matched pairs are then used to update both the spatial alignment and the time offset.
%
Closest to our work are temporal alignment methods that correlate signals derived from positions alone: \citet{wang2026lidar} cross-correlate or dynamic-time-warp speed profiles, while \citet{merkle2023automated} align a reference trajectory recorded with an \ac{RTS} to an estimated \ac{SLAM} trajectory by matching, at each timestamp, the distance to each trajectory's starting position.
Unfortunately, both solutions require sufficiently varied motion patterns to produce a distinguishable signal, failing near constant speed.
In this work, we propose a method based on the curvature signature of two trajectories, which is less sensitive to drift and less demanding on speed estimation.
\subsection{Trajectory evaluation metrics}
\label{sec:RW-C2}
Trajectory evaluation metrics evolved especially with \ac{SLAM} benchmarks, from map-based comparisons to pose-based errors, each generation evolving from the precedent one, while carrying the assumptions of its predecessors.
While before 2009, trajectory and map comparisons were presented mostly as qualitative results, the use of trajectory evaluation metrics evolved rapidly in recent years to become mandatory nowadays.
\subsubsection{Early-trajectory based SLAM evaluation (2009-2012)}
\citet{kummerle2009measuring} shifted \ac{SLAM} evaluation from comparing maps to the estimated robot trajectory during data acquisition.
They propose a protocol for analyzing the results of \ac {SLAM}- based trajectories on a metric, enabling an objective comparison of different \ac{SLAM} approaches.
However, this protocol relies on manually aligned laser scans between selected pairs of poses to obtain the true relative motion between them~\citep{kummerle2009measuring}.
Afterward, \citet{burgard2009comparison} proposed a trajectory-based benchmark for \ac{SLAM}, in which performance is measured on the corrected trajectory using relative relations between poses, rather than a global reference frame, enabling objective comparison across different estimation methods and sensor modalities, while avoiding the inability to evaluate when global pose error arises.
\subsubsection{The rise and standardization of ATE/RPE metrics (2012-2021)}
\citet{sturm2012benchmark} introduced the TUM RGB-~D benchmark, a visual SLAM benchmark, with a motion-capture reference trajectory, as well as novel evaluation metrics, the \ac{RPE} and \ac{ATE}.
These metrics became the standard for most \ac{SLAM} benchmarks, building on the relative pose error proposed by \citet{kummerle2009measuring}, which led to automatic evaluation toolkits.
The \ac{ATE} measures global consistency after alignment, while \ac{RPE} captures local motion accuracy and drift over fixed intervals~\citep{fontan2024look}.
The field of visual \ac{SLAM} consolidated around these metrics, which were originally intended for indoor motion capture systems producing high acquisition rate, well-synchronized, and accurate 6-\ac{DoF} trajectories.
\citet{geiger2012we} instead extended \citet{kummerle2009measuring} relative pose formulation in a visual \ac{SLAM} benchmark, measuring translational and rotational drift over fixed-length trajectory segments or outdoor visual odometry and \ac{SLAM} evaluation.

Later, \citet{zhang2018tutorial} clarified that quantitative trajectory evaluation is not just metric selection, but also a problem of alignment choices tied to sensing modality.
Rather than proposing a new benchmark, they showed that many details, often described vaguely in the literature, affect the final score, and that this ambiguity lowers reproducibility.
As a key result,  they demonstrated that \ac{ATE} tends to decrease when more states are used during alignment, showing that the reported error depends not only on the estimated trajectory, but also on the evaluation protocol.
They also explain that \ac{RPE} is less sensitive than \ac{ATE} to punctual errors in the estimated trajectory, and the \ac{ATE} tends to have a larger error when a punctual error happens at the beginning of the trajectory rather than at the end.
These conclusions lead to researchers reporting both metrics in further benchmarking.
\ac{ATE} and \ac{RPE} fit in the \emph{spatial measures} category as described by \citet{ranacher2014compare}, a category of similarity measures that compare \emph{spatiotemporal} positions with respect to space and neglect time, hence the need to temporally align them before evaluating.
Although the literature presents a strong adoption of trajectory metrics, benchmark datasets, and alignment choices, we observed little research on a universally accepted evaluation protocol that specifies spatial alignment, temporal alignment, and parameter tuning, which is the focus of this paper.
\subsubsection{Alternative metrics (2021-now)}
Recent papers increasingly argue that \ac{ATE} is useful but insufficient.
\citet{lee2024s} showed that the alignment algorithm required by \ac{ATE} is highly sensitive to outliers because the metric is based on an L2 norm alignment to minimize \ac{RMSE} residuals.
They proposed the Discernible Trajectory Error (DTE) and Discernible Rotation Error (DRE) as robust alternatives based on median-centered alignment, robust rotation fitting, and winsorized normalized residuals.
The advantage of these metrics is to be more sensitive to varying accuracy, but the method was only verified in simulations.
\citet{jin2021image} introduced the \ac{mAA}, a metric based solely on angular error.
The \ac{mAA} is obtained as the area under the area under the normalized cumulative histogram of the angular differences between the estimated and reference translation and rotation vectors between every possible pair of cameras~\citep {lee2025alignment}.
Then, \citet{lee2025alignment} extended the formal metric with \ac{TAS}, \ac{RAS}, and \ac{PAS}, arguing that a good metric should be robust to outliers, insensitive to trajectory length, tolerant to collinear motion (\eg common on wheeled vehicles), and not require metric-unit reference (\ie absolute scale).
In their comparison, \ac{ATE} fails the robustness criterion, \ac{mAA} is robust to outliers but not to trajectory length and collinearity, DTE improves robustness, but still depends on dataset-sensitive choices, while \ac{TAS}, \ac{RAS}, \ac{PAS} are presented as satisfying all criteria, though designed for multiview pose accuracy in \ac{SfM}.
However, DTE/DRE, \ac{mAA}, and \ac{TAS}/\ac{RAS}/\ac{PAS} all presuppose full 6-\ac{DoF} pose estimates or 3-\ac{DoF} orientations for both trajectories, and are therefore inapplicable when the reference trajectory records position only, as is the scenario for most outdoor and field robotics benchmarks.

Other work avoids this requirement entirely by removing the reference trajectory in the evaluation protocol.
\citet{recasens2023drunkard} proposed a novel ground truth-free tracking error metric, the  \ac{APTE}, that does not require reference data for their new odometry system and dataset.
They generate loop videos by duplicating and reversing a given image sequence and concatenating it to the end of the original sequence.
\citet{fontan2024look} introduced another ground truth-free metric, the \acf{GTF}-\acf{ATE}, for evaluation of \ac{SfM} and visual \ac{SLAM} systems. 
However, these metrics and evaluation methods are only applicable to videos for visual odometry.
The most common metric used when a reference trajectory only records 3-\ac{DoF} is the Euclidean distance\footnote{Also known as \emph{point distance} in the \href{https://github.com/MichaelGrupp/evo}{evo} package.} as used by \citet{frey2026grandtour}, which measures the relative displacement between the starting point and the end point between the estimated and the reference over a fixed window, as detailed by \citet{ranacher2014compare}.
In this work, we propose an evaluation metric that leverages curvature to compare trajectories from 3D positions, invariant to trajectory orientation and compensating for miscalibration or unknown extrinsic calibration.
\section{Theory}
\label{sec:theory}
The evaluation framework introduced in this paper consists of two complementary steps: (1) a curvature-based temporal alignment that estimates the time offset between the estimated and reference trajectories, and (2) a trajectory evaluation metric, \ac{DE}, that quantifies the estimation error with respect to the distance travelled.
\subsection{Curvature-based time alignment}
\label{sec:theory-C1}
Any comparison between an estimated and a reference trajectory presupposes that the two are time-calibrated, \ie that the time offset $\Delta t$ between the two sensor clocks is known.
In practice, $\Delta t$ is rarely available \textit{a priori} or could be wrongly assumed to be given, since the estimated and reference sensors are typically triggered by independent, unsynchronized clocks.
Therefore, we propose a solution to recover $\Delta t$ directly from the estimated and reference trajectories and their timestamps.

\begin{figure}[htbp]
    \centering
    \includegraphics[height=1.7in]{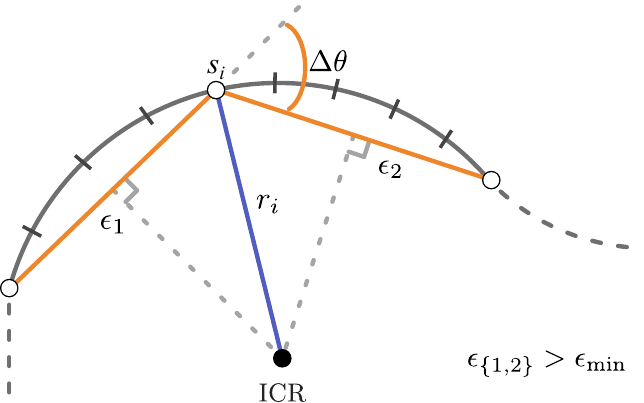}
    \caption{
    Discrete curvature estimation from three trajectory points selected using a minimum length $\epsilon_{\min}$. The turning angle $\Delta\theta$ between the two chord lengths $\epsilon_1$ and $\epsilon_2$ defines the curvature $\kappa_i$ and its associated radius $r_i$. Ticks denote trajectory samples, and the black dot indicates the \ac{ICR}. Note that $\theta$ characterizes the trajectory geometry and is independent of the platform heading (yaw angle).
    }
    \label{fig:theory-C1}
\end{figure}

Our solution relies on the fundamental property of rigid body motion that enforces that all points on that rigid body must share the same \ac{ICR} at a given time.
Consequently, the associated radius of curvature $r$ varies synchronously across all trajectories, regardless of the position of the observed point on the body.
Moreover, it is invariant to rigid transformations, making the time calibration procedure robust to wrong spatial calibration.
For each trajectory sample $i$, the curvature $\kappa_i$ is estimated over an $\epsilon_{\min}$-neighbourhood:
\begin{equation}
\kappa_i = \frac{1}{r_i} = \frac{2\Delta \theta}{\epsilon_1 + \epsilon_2},
\label{eq:curvature}
\end{equation}
where $\Delta\theta$ is the angular change subtended over chord length $\epsilon_{1}$ and $\epsilon_{2}$ as illustrated in~\cref{fig:theory-C1}.
To avoid singularities and numerical instabilities, the curvature is considered computable only if three constraints are satisfied.
First, $\epsilon_{\{1,2\}} >~\epsilon_{\min}$, with $\epsilon_{\min}\in\mathbb{R}^+$, preventing $\epsilon_1+\epsilon_2 \rightarrow 0$.
Second, we want to detect discontinuities in the trajectory that could break our space-curve assumptions.
These discontinuities correspond to a sharp turn where $\Delta\theta > \theta_{\max}$ in rad, and a stationary robot  $\Delta t_{\{1,2\}} \geq \Delta t_{\max}$ in seconds.
In both cases, the computation of $\kappa_i$ will be skipped.

After independently computing the curvature signatures of the estimated and reference trajectories, denoted by $\kappa^{\mathrm{est}}$ and $\kappa^{\mathrm{ref}}$, respectively, we define their normalized discrete cross-correlation for a candidate lag $k$ as
\begin{equation}
\rho(k) =
\frac{\displaystyle\sum_{i\in\mathcal{I}k}
\kappa_i^{\mathrm{est}}
\kappa_{i+k}^{\mathrm{ref}}
}{
\sqrt{\displaystyle\sum_{i\in\mathcal{I}k}
\left(\kappa_i^{\mathrm{est}}\right)^2
\displaystyle\sum_{i\in\mathcal{I}k}
\left(\kappa_{i+k}^{\mathrm{ref}}\right)^2
}},
\label{eq:xcorr}
\end{equation}
where $\mathcal{I}_k$ denotes the set of indices for which both curvature signatures are defined. 
The temporal offset between the two trajectories is then estimated as the lag that maximizes the correlation
\begin{equation}
k^\star = \operatorname*{arg\,max}_{k\in[-k_{max},k_{max}] }\rho(k),
\qquad
\Delta t = k^\star \delta t,
\end{equation}
where $\delta t$ is the sampling period. Both trajectories are upsampled using linear interpolation, so the offset is not bounded by the coarsest sensor rate.
The search window $[- k_{\max}, +k_{\max}]$ bounds $k$ to physically plausible clock offsets, and the global maximum of $\rho(k)$ corresponds to the lag at which the curvature signals are best aligned~\citep{knapp1976generalized}.
Once $\Delta t$ is recovered, the estimated timestamps are shifted, $t_k^{\text{est}} \leftarrow t_k^{\text{est}} + \Delta t$, and the two trajectories are treated as time-calibrated for the metrics introduced next.

\subsection{Drift Error (DE)}
\label{sec:theory-C2}

\subsubsection{Metric Definition}
To address the limitations of \ac{ATE} and \ac{RPE}, which require full 6-\ac{DoF} poses, we propose the \acf{DE}.
The metric compares the distances travelled by the estimated and reference trajectories without any spatial alignment needed. 
Requiring only 3D reference positions is particularly suitable for large-scale scenarios where scalable positioning systems (\eg \ac{GNSS} or \ac{RTS}) provide the reference trajectory.
Given two time-aligned trajectories, \ac{DE} is defined as the ratio between the travelled distances,
\begin{equation} \label{eq:DE}
    \text{DE} = \left| 1 - \frac{\Delta s^{\text{est}}}{\Delta s^{\text{ref}}} \right|,
\end{equation} 
where $\Delta s^{\text{est}}$ and $\Delta s^{\text{ref}}$ denote the path lengths travelled by the estimated and reference trajectories, respectively, over the same window length $w\in\mathbb{R^+}$ in meters.
Fixing a window length to split a trajectory into multiple observation points is standard practice for \ac{RPE} to reduce the impact of punctual events biasing the \ac{ATE} metric.
Since we measure the distance travelled along the reference to evaluate to satisfy the threshold fixed by $w$, we can avoid singularity in \cref{eq:DE} as $\Delta s^{\text{ref}} \leq w$, with the difference between both values depending on the sampling rate.
When computed over multiple windows, \ac{DE} yields a distribution that can be summarized using appropriate descriptive statistics (\eg mean, \ac{RMSE}, median, standard deviation, interquartile range).

\subsubsection{Minimizing the lever arm impact}
\begin{figure}[htbp]
    \centering
    \includegraphics[width=0.9\linewidth]{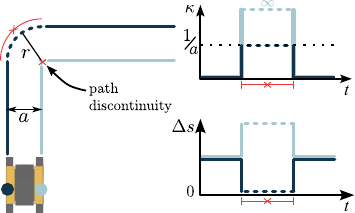}
    \caption{
    Example of a rare degenerate configuration where one point moves on a space curve while the other one does not.
   Two uncalibrated sensors, separated by a lever arm $a$, are rigidly mounted on the same platform. During the turn, the outer sensor (dark blue) traces an arc and travels a distance $s$, with a curvature radius equal to $a$, while the inner sensor (light blue) remains temporarily stationary. This degeneracy is explicitly handled by our metric.
    }
    \label{fig:theory-C2}
\end{figure}

The lever arm, defined as the translation component of the extrinsic calibration between sensors, may be inaccurate or even unknown.
Such an error can bias standard trajectory metrics (\eg \ac{ATE}, \ac{RPE}) as well as \ac{DE}, since different lever arm magnitudes result in different travelled distances during rotational motion. 
In our framework, the local curvature radius provides a direct means to correct this effect using the arc-length relation.
Reusing the curvature radius $r$ introduced in \cref{eq:curvature}, two points rigidly connected to the same rigid body undergo the same angular displacement $\Delta\theta$ while tracing concentric arcs. 
Their theoretically travelled distances $\Delta s^*$ therefore differ only through their respective radii, yielding the ratio $\alpha$:
\begin{equation}
\alpha = \frac{\Delta s^{\text{ref*}}}{\Delta s^{\text{est*}}} = \frac{r^{\text{ref}} \Delta\theta}{r^{\text{est}} \Delta\theta} = \frac{r^{\text{ref}}}{r^{\text{est}}}.
\label{eq:alpha-correction}
\end{equation}
Using this relation, \ac{DE} can be corrected for the effect of the lever arm through the following more general formulation:
\begin{equation}
    \text{DE} = \left| 1 - \alpha \frac{\Delta s^{\text{est}}}{\Delta s^{\text{ref}}} \right|.
\label{eq:DE-final}
\end{equation}
As this correction is not trivial, we examine three scenarios to explain the behaviour of the correction:
\begin{enumerate}
\renewcommand{\labelenumi}{\emph{(\roman{enumi})}}
    \item In the case of an accurate extrinsic calibration, both trajectories refer to the same point on the rigid body. 
    Hence, $r^{\text{ref}} = r^{\text{est}}$ leaving \ac{DE} unchanged with $\alpha = 1$. 
    \item In the case of straight-line motion,
    the curvature radii tend to infinity, while their difference remains bounded by the finite lever arm. 
    Consequently, their ratio tends to one, yielding $\alpha \rightarrow 1$.
    \item Finally, \cref{fig:theory-C2} illustrates the degenerate case where the robot rotates about the tracked reference point, leading to $r^{\text{ref}}=0$. 
    While the reference point remains stationary, other points on the rigid body travel along an arc, causing the correction ratio to become unbounded. Such low-radius segments are therefore detected during curvature computation and excluded from the \ac{DE} evaluation.
\end{enumerate}
As a practical note, curvature estimation may be affected by noise in real-world experiments. In theory, the difference between the two turning radii is bounded by the lever arm magnitude $a$, \ie $\Delta r = \left|r^{\text{est}}-r^{\text{ref}}\right| \in [0, a]$.
Therefore, when $\Delta r>a$, the discrepancy cannot be explained by the lever arm alone, and no lever arm correction is applied ($\alpha=1$).

\section{Results}
\label{sec:results}
Instead of simulating different trajectories, we decided to extract real trajectories publicly available to showcase different scenarios.
Therefore, we propose different case studies, instead of large-scale evaluation, to guide researchers needing to evaluate multiple algorithms.

\subsection{Case study: Impact of time calibration}
\label{sec:results-C1}

\begin{figure}[htbp]
    \centering
    \includegraphics[width=1\linewidth]{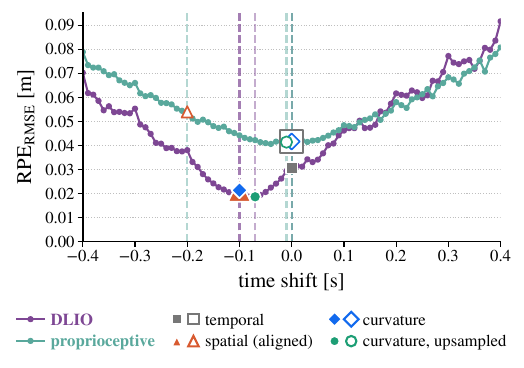}
    \caption{
    Time calibration on the GrandTour dataset HEAP-1 mission.
    The curves trace the \ac{RPE}$_\text{RMSE}$ error as a function of the time shift applied to the lidar-inertial-odometry (purple) and legged-odometry (teal) estimates.
    The markers show the offset recovered by each method, filled for the lidar-inertial-odometry and hollow for the legged-odometry. 
    A method is accurate when its marker sits at the minimum of its error curve.
    }
    \label{fig:result-C1}
\end{figure}

To illustrate the significant impact of even a small temporal misalignment and evaluate our alignment method, we compare two trajectories from the HEAP-1 mission of the publicly available GrandTour dataset~\citep{frey2026grandtour} with regard to a reference trajectory.
The first trajectory is computed based on a lidar-inertial odometry estimate, \ac{DLIO}~\citep{chen2023dlio}, while the second one is a proprioceptive state estimator~\citep{bloesch2013state}. 
Both estimated trajectories were computed independently of this paper and were evaluated against a reference trajectory provided with the dataset.
First, we generated time-shifted versions of the two estimated trajectories by applying offsets $\Delta t \in [-0.4, 0.4]$ seconds and computed the corresponding \ac{RPE}$_\text{RMSE}$. 
The window size used to split the trajectory, a parameter needed to compute the \ac{RPE}$_\text{RMSE}$ is \SI{1}{\meter}.
The original trajectories are assumed to be time-synchronized through a hardware clock (\ie \ac{PTP}) and therefore to have negligible offset with respect to the reference trajectory. 
The minimum error is thus expected at $\Delta t=$ \SI{0}{\second}.
As shown in \cref{fig:result-C1}, \ac{DLIO} (purple) generally outperforms the proprioceptive solution (teal), consistent with the original evaluation. 
However, this ranking strongly depends on temporal alignment: for sufficiently negative time shifts, the proprioceptive solution outperforms \ac{DLIO}, while for positive shifts the performance of the two methods becomes comparable and may eventually reverse. 
Thus, even a relatively small temporal misalignment can alter the relative ranking of the evaluated methods.

Moreover, we used both trajectories to evaluate three time calibration alignment methods: temporal, spatial, and curvature (ours).
The \emph{temporal} method assumes that the calibration is given (\ie $\Delta t =$ \SI{0}{\second}), which is the common practice of handling time in evaluation toolkits.
The second method, \emph{spatial}, minimizes a spatio-temporal alignment problem, pairing the estimate's timestamp with the reference's timestamps after performing an Umeyama $\SE(3)$ rigid spatial alignment.
Finally, our proposed method, \emph {curvature}, applies a cross-correlation over the curvature signature.
The results for these different alignment methods are reported as coloured markers and vertical dashed lines in \cref{fig:result-C1}.
Focusing on the proprioceptive error, we see that both the temporal and curvature solutions agree on a time shift close to zero, which correlates with the lowest reported error.
The spatial solution reports a time offset of 0.2 seconds, which does not align with the lowest error.
However, when it comes to \ac{DLIO}, both spatial and curvature methods identified a time shift of \num{-0.1} second, with the curvature upsampled solution marginally improving the time shift.
When investigating the source of this offset, we notice that the lidar used for \ac{DLIO} records at \SI{10}{\Hz}, and this sensor is not used to compute the proprioceptive trajectory.
With a time offset close to \SI{-0.1}{\second}, the error is inflated by a \SI{65}{\percent} with a reported error of \SI{3.06}{\centi\meter} instead of \SI{1.86}{\centi\meter}.


\subsection{Case study: impact of lever arm on DE evaluation metrics}
\label{sec:results-C2}
\begin{figure}[htbp]
    \centering
    \includegraphics[width=1\linewidth]{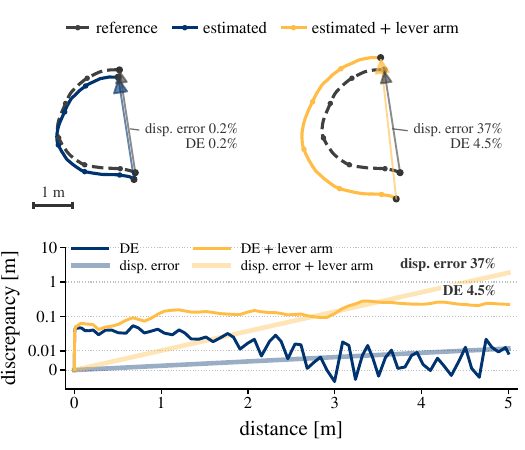}
    \caption{
    \ac{DE} evaluation metric against the state-of-the-art displacement error (Euclidean Distance) on the GrandTour EIG-2 mission.
    \emph{Top figure}: planimetric view of one window length, the lidar-inertial-odometry estimate (blue) and its copy with an injected lateral lever arm (yellow) are read against the \ac{INS} reference (dashed).
    The transparent arrows mark the start-to-end \emph{point distance} displacement error, and the plain line is the path used for the \ac{DE} error.
    \emph{Bottom figure}: distance discrepancy over the window; the path-length correction curates the bias the lever arm injects into the \ac{DE}, a bias the displacement cannot separate from estimation error.
    Note that the $y$ axis is on a log scale.
    }
    \label{fig:results-C2}
\end{figure}

We want to demonstrate the capability of \ac{DE} to be robust to errors caused by a lever arm (\ie a error in the extrinsic calibration of the sensors).
To showcase this capability, we extracted the first \SI{5}{\m} of the GrandTour EIG-2, where the robot does a large arc circle.
We selected an estimated trajectory based on \ac{DLIO} and the reference computed using an \ac{INS}.
As shown in the top-left corner of \cref{fig:results-C2}, in our nominal case, both trajectories are visually equivalent when computing 3-\ac{DoF} evaluation metrics, as both the displacement error (\ie point distance in the \href{https://github.com/MichaelGrupp/evo}{evo} package) and \ac{DE} report an error of \SI{0.2}{\percent}.
To assess the impact of a lever arm, we manually added a constant lateral offset of \SI{0.50}{\meter} in the robot frame for the estimated trajectory, which is displayed as a yellow line in the top-right corner of \cref{fig:results-C2}.
When computing both metrics on this perturbed trajectory, \ac{DE} measures a drift of \SI{4.5}{\percent}, while the displacement error is \SI{37}{\percent}.
Although we cannot say that \ac{DE} is immune to an error in the lever arm, we can show that its impact is less severe on \ac{DE}, thus limiting this potential bias in a complete evaluation.
For example, a \ac{VO} solution based on a miscalibrated camera could be wrongly ranked against a \ac{LO} solution based on the well-calibrated lidar.

\subsection{Case study: parameter sensitivity}
\label{sec:results-C3}
For this study, we evaluated the impact of parameter selection over multiple trajectories in two different applications.
We selected nine trajectories from the GrandTour dataset (\ie ARC-2, CON-4, EIG-1, EIG-2, GRI-1, HEAP-1, SBB-1, SNOW-2, and SPX-2) and also nine trajectories from the FoMo dataset~\citep{boxan2026fomo} (\ie the orange section repeated through a year in different weather conditions). 
Key differences between these two datasets are listed in \cref{tab:params-C3}.
The two parameters related to the space-curve discontinuities are $\Delta t_\text{max} = 2$~seconds and $\theta_\text{max}=\pi / 2$~rad.
In terms of methodology, for each parameter studied, we freeze the rest of the parameters to a constant, also listed in the same table.
This frozen parameter is also in \textbf{bold} in all the subsequent plots.
For a given trajectory, we split the reference trajectory into $n$ windows of length $w$ and allow these windows to overlap by $\beta$~\si{\%}.
The proposed metric \ac{DE} is computed for each window.
At this point, we compute the \ac{RMSE} over $n$ observations for a given trajectory.
To validate that a given parameter is stable over all the trajectories, we plot the evolution of the median and \ac{IQR} over the evaluated parameter. 

\begin{table}[htbp]
    \centering
    \caption{Differences between the two datasets used in this case study.
    \emph{Top section}: key characteristics of the recorded data.
    \emph{Middle section}: evaluation parameters proposed for \ac{DE}.
    \emph{Bottom section}: evaluation parameters generic for window-based metrics (\eg \ac{RPE}, displacement error).
    }
    \label{tab:params-C3}
    \begin{tabularx}{\linewidth}{@{} X l l @{}}
    \toprule
         &  \textbf{GrandTour}~\citep{frey2026grandtour} & \textbf{FoMo}~\citep{boxan2026fomo} \\
    \midrule
    Reference traj. source                 & \ac{INS} & Differential \ac{GNSS} \\
    Reference traj. uncertainty (est.)     & \SI{0.03}{\m} & \SI{0.1}{\m} \\
    Reference traj. \ac{DoF}                      & 6 & 3 \\
    Estimated traj. source                  & Leg odometry & Wheel odometry \\
    Robot maximum speed              & \SI{0.5}{\meter\per\second} & \SI{2.5}{\meter\per\second} \\
    Locomotion type                  & legs & wheels and tracks \\
    Robot footprint (diagonal)       & \SI{0.8}{\m} & \SI{1.5}{\m} \\
    Trajectory length (mean)       & \SI{214}{\m} & \SI{2200}{\m} \\
    \midrule
    Minimum displacement $\epsilon_\text{min}$  & \SI{0.1}{\m} & \SI{0.3}{\m} \\
    Expected lever arm $a$                      & \SI{0.1}{\m} & \SI{0.1}{\m}\\
    \midrule
    Window length $w$                           & \SI{5}{\m} & \SI{5}{\m}\\
    Window overlap $\beta$                     & \SI{0}{\percent} & \SI{0}{\percent} \\
    \bottomrule
    \end{tabularx}
    \label{tab:placeholder}
\end{table}

\begin{figure}[htbp]
    \centering
    \includegraphics[width=\linewidth]{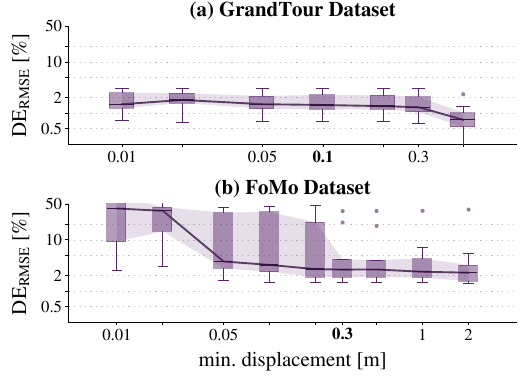}
    \caption{
    Sensitivity evaluation of the minimum displacement $\epsilon_{\text{min}}$.  
    A bold tick indicates the parameter held constant during the other evaluation.
    }
    \label{fig:results-C3-A}
\end{figure}

First, we evaluate the impact of the minimum displacement $\epsilon_\text{min}$ parameter.
This parameter is used in the computation of the curvature and acts as a low-pass filter to counteract the noise in the reference trajectory.
Setting $\epsilon_\text{min}$ too small will create high curvature following the noise, while too high will flatten any turns in the trajectory.
Furthermore, the longer the distance used for $\epsilon_\text{min}$ is, the more points on the trajectory will be skipped, leading to very few points to evaluate the error. 
\Cref{fig:results-C3-A} shows these relations for $\epsilon_\text{min}$ in the two datasets under study.
For both datasets, we observe that the larger $\epsilon_\text{min}$ is, the lower \ac{DE} becomes.
Also, above \SI{0.5}{\m}, many points start to be discarded (\ie see the gray dashed line rising), leading to an evaluation based on too few points.
One should note that, even at low $\epsilon_\text{min}$, there is already \SI{20}{\%} of the points discarded in GrandTour compared to less than \SI{1}{\%} for FoMo.
This difference comes from the legged robot being more stationary during each trajectory compared to the ground vehicle used in FoMo.
This filtering is an inherent property of converting the evaluation over a space curve instead of a trajectory, as a robot staying static for a very long time could produce very low error and bias an evaluation.
Focusing on the FoMo dataset, which has a higher noise level on the reference trajectory, we see that setting $\epsilon_\text{min}$ lower than \SI{0.05}{\m} causes a rapid rise in the error, which is not the case for GrandTour.
At least for the FoMo dataset, even if we had only access to a coarse estimate of the reference uncertainty, we see a plateau in \cref{fig:results-C3-A} indicating that the parameters do not affect the error between \SI{0.05}{\m} and \SI{0.5}{\m}.
Therefore, even if we only have two noise levels to corroborate our observations, we propose that a reasonable guess for $\epsilon_\text{min}$ is three times the expected standard deviation of a reference trajectory.


\begin{figure}[htbp]
    \centering
    \includegraphics[width=\linewidth]{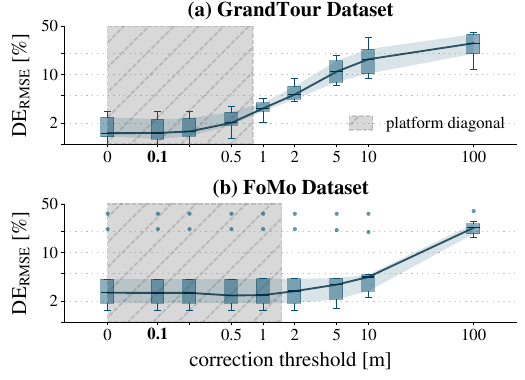}
    \caption{
    Sensitivity evaluation of the expected lever arm $a$.  
    A bold tick indicates the parameter held constant during the other evaluation.
    The gray background marks the threshold below each platform's own footprint (\ie the platform diagonal).
    }
    \label{fig:results-C3-B}
\end{figure}

Second, we evaluate the impact of an expected error in the extrinsic calibration by varying the expected lever arm $a$ parameter.
As this is a bounding value, we expect little impact on the error if this lever arm is within reasonable values.
Moreover, the two datasets used for this evaluation were curated, and an extensive amount of time was spent in the calibration process.
Therefore, we should expect a low value for $a$, which we can observe in \cref{fig:results-C3-B}.
We see that the impact of this parameter has a plateau at zero for both datasets and then rises around \SI{0.5}{\m} for GrandTour and \SI{5}{\m} for FoMo, giving us some room to select this parameter without impacting the error too much.
As a reminder, the threshold distance $a$ is used to make the distinction between error coming from poor performing algorithm and an error caused by poor calibration for part of the trajectory with high curvature.
In the case of a well-calibrated dataset, $a$ can be set to a low value without influencing the error while allowing potential calibration to be mitigated.
For this reason, we proposed $a=0.1$~meter for both datasets.
In the case of an evaluation between two trajectories where the calibration is unavailable, a coarse estimate for the expected lever arm $a$ is the longest possible distance between two sensors.


\begin{figure}[htbp]
    \centering
    \includegraphics[width=\linewidth]{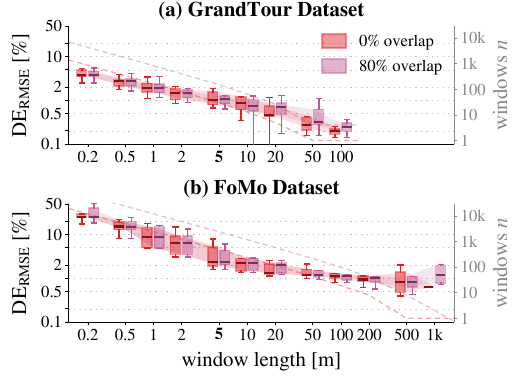}
    \caption{
    Sensitivity evaluation of the window length $w$ with disjoint and overlapping windows. 
    A bold tick indicates the parameter held constant during the other evaluation.
    The colored dashed line is a secondary axis showing the number of windows $n$.
    }
    \label{fig:results-C3-C}
\end{figure}

Finally, unlike the previous parameter studies, the window length $w$ and the window overlap $\beta$ used to split a trajectory do not directly impact the error in the process and are not specific to \ac{DE}.
The window length $w$ changes how the error is reported.
For example, a \ac{DE} value of 0.03 evaluated over multiple windows of \SI{5}{\m} length would read: The solution X is expected to drift by \SI{3}{\%} for every \SI{5}{\m} driven.
The overlap parameter $\beta$ increases the number of possible windows $n$ where $\beta=0$ would give disjoint windows and therefore the lowest number of windows.
\Cref{fig:results-C3-C} shows the results for different window lengths and two extreme values of overlap.
The first key observation is that a large overlap $\beta$ leads to the same error and artificially augments the number of observations $x$.
Indeed, the larger the overlap is, the more correlated the information becomes as the same part of the trajectory is reused many times.
This practice is often used when trajectories are too short, leading to too few observations.
An interesting observation is that the error appears to plateau beyond \SI{20}{\m} for the GrandTour dataset and beyond \SI{200}{\m} for the FoMo dataset, leading to the conclusion that in general their odometry algorithm is expected to drift by \SI{0.5}{\%} and \SI{1}{\%}, respectively.
Therefore, we encourage the use of only disjoint windows (\ie $\beta=0$~\%) to avoid biasing the number of real observations.
Another point is that setting the window length $w$ too low will highlight noise in the reference trajectory more than the behavior of the reference.
For a high window length $w$, we are confident that we observe the drift of the estimated reference, but we reduce the number of observations heavily, thus being more sensitive to outliers.
We suggest setting the window length $w$ while keeping enough meaningful observation points.
Once fixed, the same window length $w$ should be used across solutions, trajectories, and datasets to ensure a fair comparison.


\section{Conclusion}
\label{sec:conclusion}

This work revisited trajectory evaluation for datasets that provide a reference with 3-\ac{DoF} positions only, where the standard metrics, \ac{ATE} and \ac{RPE}, are undefined.
The proposed protocol using \acl{DE} can mitigate the impact of wrong time and extrinsic calibrations, therefore opening evaluations on less curated datasets.
We showed that our curvature-based calibration method recovers a time offset from an estimator using time-synchronized sensors, a small offset of \SI{70}{\milli\second} inflating the error by \SI{65}{\percent}.
Demonstrating that if the time offset is left uncorrected, the ranking in the evaluation process can be altered and impact algorithm performance in benchmarks.
Our proposed evaluation metric \acf{DE} can reduce the lever arm impact from missing extrinsic calibration information, where we added a manual \SI{0.5}{\meter} lateral offset, reporting \SI{37}{\percent} for the state-of-the-art displacement error for 3-\ac{DoF} positions compared to \SI{4.5}{\percent} for our metric.
We analyzed the sensitivity of the parameters for the curvature computation and \ac{DE} metric using the GrandTour dataset~\citep{frey2026grandtour} and the FoMo dataset~\citep{boxan2026fomo}, standardizing the evaluation protocol.
%
Both contributions require curvature excitation and degrade on straight-line segments.
Future work aims for curvature and torsion estimation directly on the full 3D curve, an orientation metric, as well as an analysis of the different references' bias into the proposed evaluation metric.

\section*{ACKNOWLEDGMENT}

Generative AI (Claude, Anthropic) was used to assist in generating the plotting code for the data-driven result figures in~\cref{sec:results}; the underlying data, analysis, and figure content were produced and verified by the authors.
\printbibliography
\end{document}